\documentclass[10pt,twocolumn,letterpaper]{article}

\usepackage{iccv}
\usepackage{times}
\usepackage{epsfig}
\usepackage{graphicx}
\usepackage{amsmath}
\usepackage{amssymb}

\usepackage{breqn}
\usepackage{siunitx}
\usepackage{cite}
\usepackage{ctable}
\usepackage{hhline}
\usepackage{booktabs}
\usepackage{subcaption}
\usepackage{csquotes}
\usepackage{bm}

\usepackage{framed,multirow}

\usepackage{booktabs}
\usepackage[ruled, vlined, linesnumbered]{algorithm2e}
\usepackage{algpseudocode}

\newcolumntype{L}[1]{>{\raggedright\let\newline\\\arraybackslash\hspace{0pt}}m{#1}}
\newcolumntype{C}[1]{>{\centering\let\newline\\\arraybackslash\hspace{0pt}}m{#1}}
\newcolumntype{R}[1]{>{\raggedleft\let\newline\\\arraybackslash\hspace{0pt}}m{#1}}

\usepackage[pagebackref=true,breaklinks=true,letterpaper=true,colorlinks,bookmarks=false]{hyperref}

\iccvfinalcopy 

\def\iccvPaperID{3802} 

\ificcvfinal\fi
\begin{document}

\title{PoseLifter: Absolute 3D human pose lifting network from a single noisy 2D human pose}

\author{Ju Yong Chang\\
Kwangwoon University\\
{\tt\small jychang@kw.ac.kr}
\and
Gyeongsik Moon\\
Seoul National University\\
{\tt\small mks0601@snu.ac.kr}
\and
Kyoung Mu Lee\\
Seoul National University\\
{\tt\small kyoungmu@snu.ac.kr}
}

\maketitle

\begin{abstract}
This study presents a new network (i.e., PoseLifter) that can lift a 2D human pose to an absolute 3D pose in a camera coordinate system. The proposed network estimates the absolute 3D location of a target subject and generates an improved 3D relative pose estimation compared with existing pose-lifting methods. Using the PoseLifter with a 2D pose estimator in a cascade fashion can estimate a 3D human pose from a single RGB image. In this case, we empirically prove that using realistic 2D poses synthesized with the real error distribution of 2D body joints considerably improves the performance of our PoseLifter. The proposed method is applied to public datasets to achieve state-of-the-art 2D-to-3D pose lifting and 3D human pose estimation.
\end{abstract}

\section{Introduction}
\label{sec1}

\begin{figure}
\centering
\includegraphics[width=1.00\linewidth]{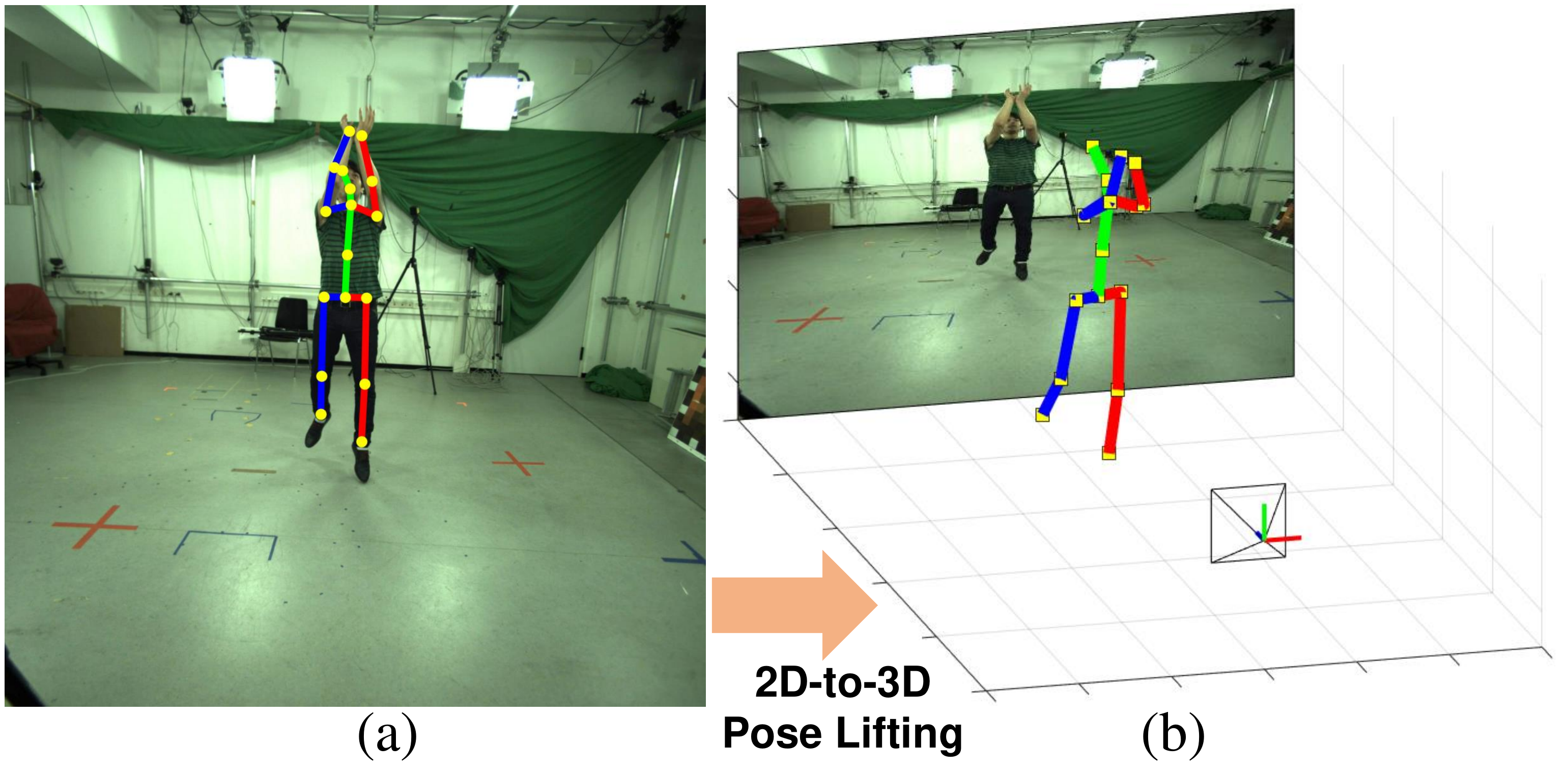}
\caption{(a) A 2D human pose consisting of a set of several joints is overlaid on a RGB image. (b) Our goal is to estimate the 3D human pose in the camera coordinate system from such sparse 2D joint information in (a).}
\label{fig1}
\end{figure}

What information can we acquire from the sparse semantic points of a single image? Figure~\ref{fig1}(a) shows a 2D human pose that consists of a set of 2D joints. Human activities can be easily recognized from such sparse 2D joint information~\cite{Johansson1975}, based on which automated algorithms~\cite{Chang2016,Jhuang2013} have been proposed. What about the 3D human pose shown in Figure~\ref{fig1}(b)? How accurately can a 3D human pose (\textit{i.e.}, 3D joint coordinates) be reconstructed using only projected geometric information without appearance features? This ill-posed problem~\cite{Lee1985}, namely, the automatic lifting of 2D joint coordinates in a single image to 3D space, has been addressed in previous studies, and successful methods have been proposed~\cite{Martinez2017,Ramakrishna2012}. However, existing methods generate only the \textit{relative 3D pose}, \textit{i.e.}, the 3D joint coordinates after the translation transform is applied to move the reference joint to the origin. To address this issue, we propose a novel 2D-to-3D pose-lifting network (\textit{i.e., PoseLifter}) that can produce an \textit{absolute 3D pose} in which the coordinates of all joints are defined on a camera coordinate system.

\begin{figure*}[t]
\centering
\includegraphics[width=1.00\textwidth]{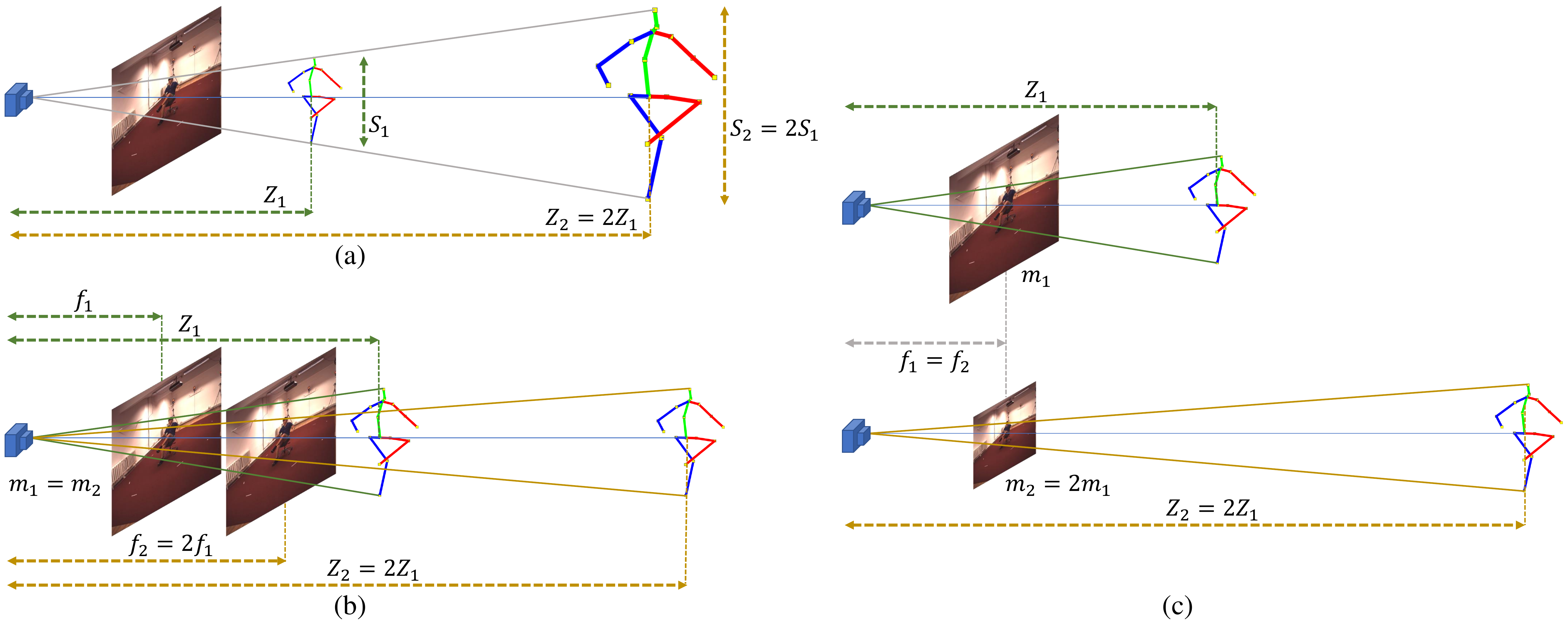}
\caption{For a given input RGB image, the root's absolute depth is proportional to the size of the human subject, as shown in (a). Even if the size of the human subject is fixed, the ambiguity in determining the absolute root depth remains due to the focal length. The focal length in terms of pixel dimensions ($\alpha$) is the product of the focal length in terms of physical dimensions ($f$) and the number of pixels per unit distance ($m$): $\alpha=f\times{}m$. Under the assumption of fixed subject size, the absolute root depth corresponding to a given input RGB image is proportional to $f$ and $m$, shown in (b) and (c), respectively. Note that when $m$ is doubled ($m_{2}=2m_{1}$), the actual physical size occupied by the image in the image sensor is halved, which doubles the absolute root depth ($Z_{2}=2Z_{1}$).}
\label{fig2}
\end{figure*}

An absolute pose can be decomposed into the absolute coordinates of a reference joint (\textit{i.e.}, the root) and the relative pose to the root. To obtain an absolute 3D pose, we use the normalized 2D pose, 2D location, and 2D scale obtained by decomposing the input 2D pose. From the obtained 2D information, location and scale enable the calculation of the root's 3D coordinates. Specifically, location and scale provide approximate information about the $(X, Y)$ and $Z$ (\textit{i.e.}, depth) coordinates of the target human subject in the camera coordinate system, respectively. The normalized 2D pose can also be used to estimate the root-relative 3D pose as in existing methods~\cite{Martinez2017}.

In this study, we propose to use all of the decomposed 2D information for the estimation of the absolute 3D pose (\textit{i.e.}, the root coordinates and relative 3D pose). This leads to the following advantages. First, under perspective projection, the distance from the camera to the human subject (\textit{i.e.}, root depth) is proportional to the ratio of the human scale in real space to the scale in the 2D image. The real scale can vary greatly depending on the posture of the human subject. For example, the scale in the squatting posture is relatively small compared to the stretching posture. This variation adds difficulty in determining the depth from only the 2D scale. However, a normalized 2D pose can provide additional information about the real scale, which helps in an accurate calculation of the depth of the root.

Secondly, determination of the root-relative 3D pose can also be aided by the 2D location and scale information. This is because the additional 2D information can alleviate the ill-posedness of the problem. For example, under the perspective projection assumption, the 2D projection of a human subject with a fixed root-relative 3D pose changes with the subject's 3D location. This further deepens the ambiguities in estimating the correct root-relative 3D pose from a given 2D pose. However, 2D location and scale provide approximate information about the subject's 3D location, thus mitigating the ambiguities of root-relative 3D pose estimation. Therefore, we propose a method of \textit{estimating root coordinates and root-relative 3D pose simultaneously using all of the normalized 2D pose, 2D location, and 2D scale information}.

Between the two goals, determining the absolute depth of the root is a significantly unconstrained problem. Technically, the ambiguities of the human subject size and camera focal length do not allow the absolute depth of the root to be uniquely determined, as illustrated in Figure~\ref{fig2}. Specifically, the absolute depth of root is proportional to both the size of human subject $S$ and the camera focal length $\alpha=f\times{}m$, in which $f$ and $m$ are the focal length in terms of physical dimension and the number of pixels per unit distance, respectively. These ambiguities are solved in this study as follows. First, in the proposed PoseLifter, the size of the human subject is learned implicitly from datasets, which resolves the \textit{size ambiguity}. To handle the \textit{focal length ambiguity}, PoseLifter outputs the \textit{canonical root depth} normalized by the focal length instead of the real depth. If additional focal length information is available, then the root's real depth can be obtained from the canonical depth.

The proposed PoseLifter can be applied to estimate the absolute 3D pose of a person from a single 2D image. A simple approach to achieve this goal is to sequentially combine a 2D human pose estimator and the PoseLifter. Specifically, the 2D human pose estimator will generate a 2D pose from the input 2D image. The resulting 2D pose will then be fed into the PoseLifter to generate the corresponding absolute 3D pose. However, the results of 2D pose estimation are unreliable because of inevitable errors. A recent study~\cite{Ronchi2017} indicated that such errors exhibit a similar distribution regardless of the type of 2D pose estimator used. Thus, a method for improving the performance of existing 2D pose estimators by utilizing such distribution was proposed~\cite{Moon2019}. In this regard, we propose to analyze the error of a 2D pose estimator and synthesize the input of the PoseLifter for learning following the resulting error statistics.

In summary, we propose the following technical improvements to 3D human pose estimation.
\begin{itemize}
  \item The first is the implementation of the \textit{normalization layer}, which is the first layer of the proposed PoseLifter. Our novel normalization layer normalizes the input 2D pose and adds the 2D location and scale information of the target subject as intermediate features. These added features enable the estimation of the root's absolute 3D coordinates and considerably improve the performance of root-relative 3D pose estimation.
  \item The second is the \textit{canonical root depth} that is independent of the camera focal length. This new representation allows the PoseLifter to be applied to any test image with an unknown focal length.
  \item The last one focuses on the \textit{connection between 2D pose estimation and 3D pose-lifting modules}. In our method, the error of a 2D pose estimator is realistically synthesized, and the result is used to learn the 3D pose lifter, thereby making the PoseLifter robust to 2D pose estimation errors. Consequently, the proposed approach achieves state-of-the-art performance on two large-scale 3D human pose datasets, namely, Human3.6M~\cite{Ionescu2014} and MPI-INF-3DHP~\cite{Mehta2017}.
\end{itemize}

The remainder of this paper is organized as follows. Section~\ref{sec2} introduces related studies. Section~\ref{sec3} describes the proposed 2D-to-3D pose-lifting method (\textit{i.e.}, PoseLifter). Section~\ref{sec4} explains the application of this method to 3D human pose estimation from a single RGB image. Section~\ref{sec5} presents the experimental results. Section~\ref{sec6} provides the conclusions drawn from the study.

\section{Related work}
\label{sec2}

\subsection{2D-to-3D human pose lifting}

In earlier studies, 3D human pose is typically modeled as a linear combination of sparse basis poses. In general, the re-projection error between the 2D projections of 3D joints and input 2D joints is minimized using optimization methods, such as greedy orthogonal matching pursuit~\cite{Ramakrishna2012}, alternating direction~\cite{Wang2014}, and convex relaxation-based alternating direction method of multipliers~\cite{Zhou2015}, along with a prior model of physically possible 3D poses~\cite{Akhter2015}. After optimization, the 3D joints in a world coordinate system and the viewpoint information of an orthographic camera are obtained simultaneously.

In recent studies, a large 2D/3D dataset was used to learn the regression function that directly converts the input 2D to the output 3D pose based on a camera coordinate system. A neural network is typically adopted for this purpose. The normalized 2D joint coordinates~\cite{Martinez2017} and a Euclidean distance matrix~\cite{Moreno-Noguer2017} are proposed as inputs for the network. All the aforementioned methods yield a relative pose based on the root, whereas our proposed method allows the acquisition of an absolute 3D pose.

\subsection{3D human pose estimation from a single image}

Recent deep learning-based methods can be divided into direct image-to-pose estimation and cascade approaches. The \textit{direct approach} produces an output 3D pose directly from an input single RGB image. These outputs exhibit various forms, such as 3D joint coordinates~\cite{Li2014}, bone-based representations~\cite{Sun2017}, and volumetric heatmaps~\cite{Pavlakos2017}. Recent studies~\cite{Moon2019b,Sarandi2018,Sun2018} proposed methods that directly regress 3D coordinates while maintaining the advantages of a volumetric heatmap representation through a soft-argmax~\cite{Luvizon2017,Nibali2018} operation. In particular, the methods presented in~\cite{Moon2019b,Sarandi2018} are related to our method, given that the approaches calculate the absolute 3D coordinates of the root. The previous methods differs from our method, which uses only 2D joint information for pose lifting, whereas the methods of~\cite{Moon2019b,Sarandi2018} require an RGB image as input. Absolute root coordinates were also computed in~\cite{Mehta2017}. To do so, the method in~\cite{Mehta2017} initially computes the root-relative pose using the network learned through transfer learning and then estimates the absolute location of the root through a post-processing step based on a closed-form formula. In~\cite{Mehta2017,Sarandi2018}, only the evaluation of the root-relative pose is performed, and the performance of absolute root location estimation is not reported.

The \textit{cascade approach} consists of two steps: (1) creating a 2D pose through a 2D pose estimator and (2) lifting this 2D pose to output a 3D pose. The acquisition of a 3D pose from a 2D pose is accomplished via neural network-based regression~\cite{Fang2018,Martinez2017,Moreno-Noguer2017} or by fitting a 3D morphable model to 2D joints~\cite{Bogo2016}. The cascade approach is flexible and easy to use owing to its modular design. Different types of datasets can also be used for learning 2D pose estimating and 2D-to-3D pose-lifting modules. Meanwhile, in the direct approach, the entire process of generating an output 3D pose from an input RGB image is optimized in an end-to-end fashion by a single cost function, which is considered to bring relatively high performance. Our method is a regression-based cascade approach, which nevertheless achieves better 3D human pose estimation performance than most existing direct approaches.

\begin{figure*}[t]
\centering
\includegraphics[width=0.95\textwidth]{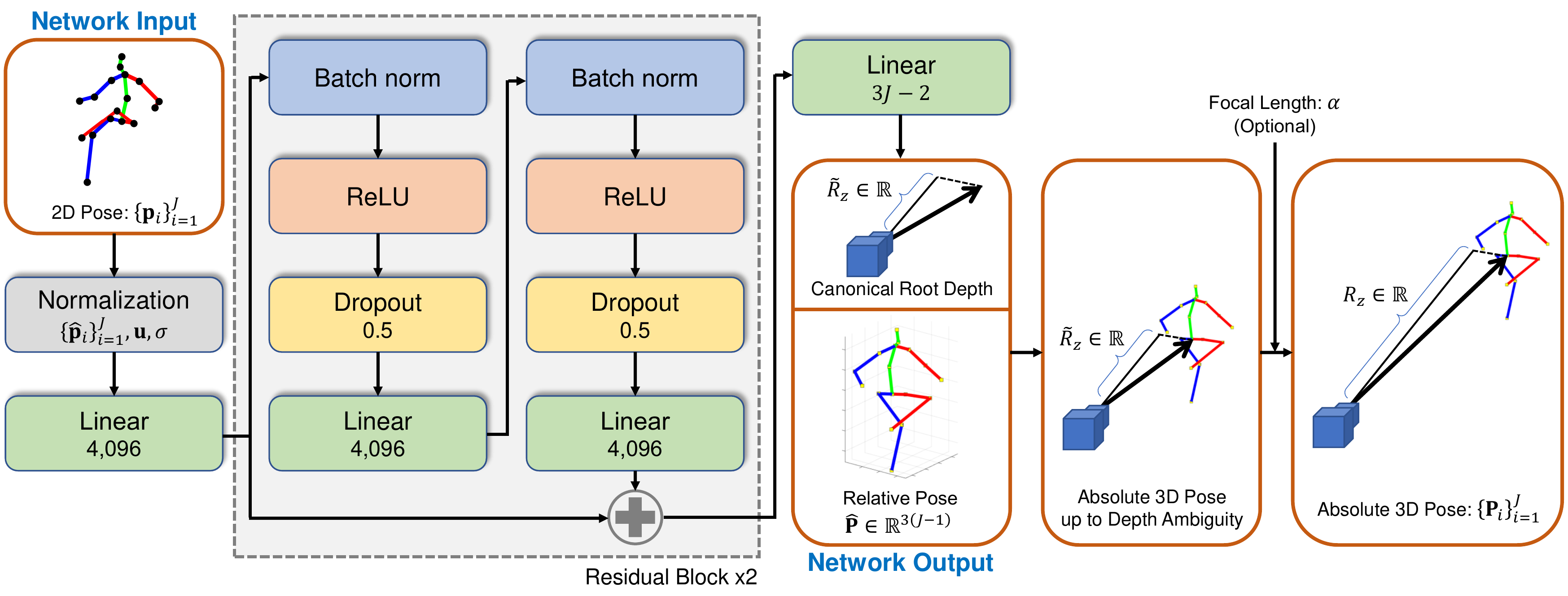}
\caption{Structure of the proposed PoseLifter.}
\label{fig3}
\end{figure*}

\subsection{Error analysis of 2D human pose estimation}

In~\cite{Ronchi2017}, prediction errors in 2D human pose estimation were categorized into various types, thereby experimentally proving that existing state-of-the-art methods produce similar error distributions. On the basis of taxonomy, a joint with a small error from ground truth is considered \textit{good}. The error of a joint that is near the ground truth but larger is called a \textit{jitter}. \textit{Inversion} and \textit{swap} represent errors due to confusion with semantically similar joints belonging to the same and different persons, respectively. Lastly, a \textit{miss} corresponds to a considerably large error that does not belong to previous cases. The method proposed in~\cite{Moon2019} refines the prediction results of existing 2D pose estimators by using estimation error statistics. In contrast to this previous study that uses a simple empirical ratio of error types, our method estimates a specific error distribution to make 3D pose lifting robust.

\section{PoseLifter}
\label{sec3}

\subsection{Input and output for PoseLifter}
\label{sec3.1}

The objective of the pose-lifting problem is to lift a given 2D pose $\{\mathbf{p}_{i}\}_{i=1}^{J}$ to a 3D pose $\{\mathbf{P}_{i}\}_{i=1}^{J}$. $\mathbf{p}_{i}\in{\mathbb{R}^2}$ and $\mathbf{P}_{i}\in{\mathbb{R}^3}$ represent the locations of the \textit{i}th joint in an input image coordinate system and a camera coordinate system, respectively. $J$ denotes the number of joints. First, we apply the two-step normalization procedure to the input 2D pose. In the first step, we convert 2D coordinates $\mathbf{p}_{i}=[x_{i},y_{i}]$ to $\mathbf{p}'_{i}=[x'_{i},y'_{i}]$ for all joints $i$ as follows:
\begin{equation}\label{eq1}
\begin{split}
x'_{i}=x_{i}-c_{x}; \\
y'_{i}=y_{i}-c_{y},
\end{split}
\end{equation}
where $(c_{x},c_{y})$ denotes the principal point of the camera. This transformation makes the coordinates of the 2D joint independent of the principal point of the camera. If the principal point of a test image is unknown, then the approximate image center can be used. According to our experiments on Human3.6M and MPI-INF-3DHP, this approximation does not yield a quantitatively significant performance difference.

In the second step, we follow the previous pose-lifting methods~\cite{Martinez2017} and further normalize the input 2D joints to zero mean and unit variance as follows:
\begin{equation}\label{eq2}
\hat{\mathbf{p}}_{i}=\frac{(\mathbf{p}'_{i}-\mathbf{u})}{\sigma},
\end{equation}
where $\mathbf{u}$ and $\sigma$ denote the mean vector and standard deviation, respectively:
\begin{equation}\label{eq3}
\mathbf{u}=\sum_{i=1}^{J}\mathbf{p}'_{i}/J,
\end{equation}
\begin{equation}\label{eq4}
\sigma=\sqrt{\sum_{i=1}^{J}\|\mathbf{p}'_{i}-\mathbf{u}\|^{2}_{2}/J}.
\end{equation}
Note that $\mathbf{u}$ and $\sigma$ represent the approximated 2D \textit{location} and \textit{scale} of the subject in the image, respectively. Under a perspective projection assumption, the 2D location and scale information provide a clue to the 3D location of the subject and allow the estimation of an absolute 3D pose. Therefore, we define a normalization layer that transforms the input $(2J)$-dimensional vector $\mathbf{p}=[\mathbf{p}_{1},\ldots,\mathbf{p}_{J}]$ into a $(2J+3)$-dimensional vector concatenated with normalized 2D coordinates, mean vector, and standard deviation. We then set this layer as the first layer of our PoseLifter. Notably, the method of~\cite{Martinez2017} does not perform normalization through the principal point, nor does it use the location and scale of the target subject.

Our objective in pose lifting is to obtain an absolute 3D pose $\{\mathbf{P}_{i}\}_{i=1}^{J}$. It can be decomposed into the root's absolute coordinates $\mathbf{R}=[R_{x},R_{y},R_{z}]\in{\mathbb{R}^{3}}$ and the relative 3D pose $\{\hat{\mathbf{P}}_{i}\}_{i=1}^{J}$ to the root: $\mathbf{P}_{i}=\mathbf{R}+\hat{\mathbf{P}}_{i}$. We estimate $\tilde{R}_{z}=\frac{R_{z}}{\alpha}$ obtained by dividing the $z$-component $R_{z}$ of the root via the focal length $\alpha$ instead of its absolute 3D coordinates. $\tilde{R}_{z}$ means a depth value that is independent of the focal length of the camera and is thus named \textit{canonical root depth} in this study. Therefore, 2D-to-3D pose lifting can be formulated as the problem of finding a \textit{3D regression function} $h:\mathbb{R}^{2J}\xrightarrow{}\mathbb{R}^{3J-2}$ that maps the input 2D pose $\mathbf{p}\in{\mathbb{R}^{2J}}$ into the canonical root depth $\tilde{R}_{z}\in{\mathbb{R}}$ and the root-relative 3D pose $\hat{\mathbf{P}}\in{\mathbb{R}^{3J-3}}$.

\begin{figure*}[t]
\centering
\includegraphics[width=0.80\textwidth]{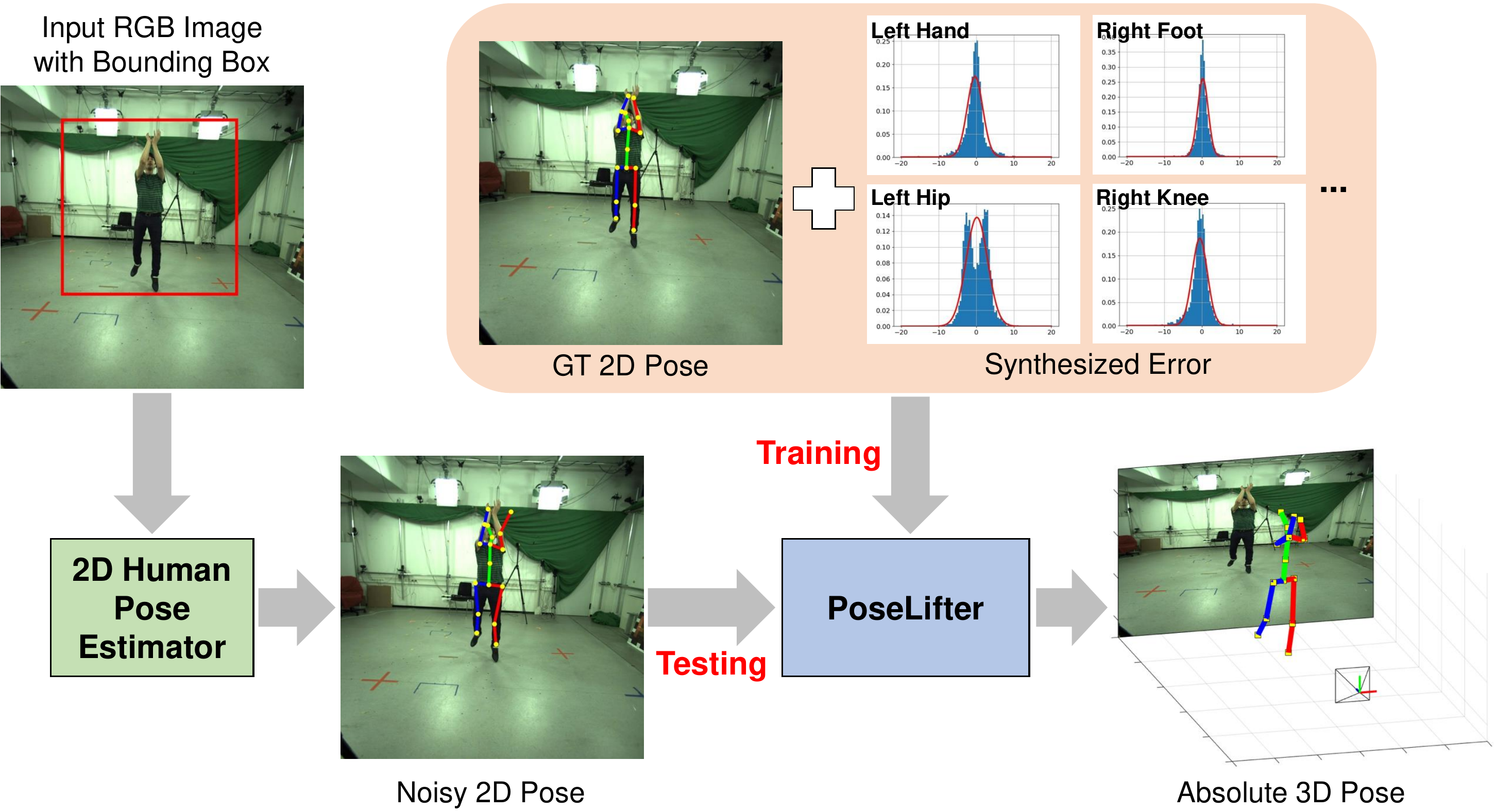}
\caption{Overview of the proposed 3D human pose estimation method from a single RGB image.}
\label{fig4}
\end{figure*}

The root with absolute coordinates $\mathbf{R}=[R_{x},R_{y},R_{z}]$ is projected to image coordinates $\mathbf{r}=[r_{x},r_{y}]$ through a camera with focal length $\alpha$ and principal point $(c_{x},c_{y})$ as follows:
\begin{equation}\label{eq4.5}
\begin{gathered}
r_{x}=\alpha\frac{R_{x}}{R_{z}}+c_{x}; \\
r_{y}=\alpha\frac{R_{y}}{R_{z}}+c_{y},
\end{gathered}
\end{equation}
which can be rearranged as follows:
\begin{equation}\label{eq5}
\begin{gathered}
R_{x}=(r_{x}-c_{x})\frac{R_{z}}{\alpha}=(r_{x}-c_{x})\tilde{R}_{z}; \\
R_{y}=(r_{y}-c_{y})\frac{R_{z}}{\alpha}=(r_{y}-c_{y})\tilde{R}_{z}.
\end{gathered}
\end{equation}
Thus the $x$, $y$-components $R_{x}$ and $R_{y}$ of the absolute root coordinates are determined by the canonical depth $\tilde{R}_z$ and the root's 2D coordinates $\mathbf{r}=[r_{x},r_{y}]$. Note that the absolute $x$ and $y$-coordinates of the root can be calculated without focal length information. Meanwhile, the absolute root depth $R_{z}$ is obtained as follows:
\begin{equation}\label{eq6}
R_{z}=\alpha\tilde{R}_{z}.
\end{equation}
Therefore, for any test image, we can reconstruct up to the canonical root depth $\tilde{R}_{z}$, which can be promoted to the absolute root depth $R_{z}$ with the help of focal length information.

\subsection{Network structure}
\label{sec3.2}

We describe the structure of our PoseLifter that realizes the 3D regression function $h$, which is shown in Figure~\ref{fig3}. Our network is primarily based on the residual block proposed in~\cite{Martinez2017}. This block iteratively passes the input vector to batch normalization~\cite{Ioffe2015}, dropout~\cite{Srivastava2014}, rectified linear unit~\cite{Nair2010}, and linear layers twice, and then adds the result to the residual connection~\cite{He2016} output. The dropout probability and feature dimension are set to 0.5 and 4096, respectively.

First, the input $(2J)$-dimensional vector is normalized to a $(2J+3)$-dimensional vector via the normalization layer. The latter is then converted to a 4096-dimensional feature vector through a linear layer. This vector passes through two residual blocks and finally outputs a $(3J-2)$-dimensional vector via a linear layer. The first number represents the canonical root depth $\tilde{R}_{z}$, while the following $3(J-1)$ numbers represent the root-relative pose vector $\hat{\mathbf{P}}$, except for the root.

Subsequently, we supervise our PoseLifter by using the ground truth root depth ${R}_{z}^{*}$, focal length $\alpha$, and relative pose $\hat{\mathbf{P}}^{*}$ for learning the 3D regression function $h$. In particular, the following cost function based on L1 loss is minimized:
\begin{equation}\label{eq7}
L=\frac{1}{N}\sum_{i=1}^{N}|\tilde{R}_{z}^{(i)} - \frac{{R}_{z}^{*(i)}}{\alpha}| + \lambda \frac{1}{N}\sum_{i=1}^{N}\|\hat{\mathbf{P}}^{(i)}-\hat{\mathbf{P}}^{*(i)}\|_{1},
\end{equation}
where superscript $i$ is the index of the sample, and $N$ denotes the total number of training samples. The first term causes our network to output a canonical root depth that is independent of the focal length. $\lambda$ is a parameter for adjusting the relative strength between the two loss functions for the canonical root depth and the root-relative pose. This parameter is set to $10^3$ in all our experiments.

\section{Cascading with 2D pose estimator}
\label{sec4}

An absolute 3D pose can be obtained from a single RGB image by using the PoseLifter presented in the previous section. The basic concept is to combine a state-of-the-art 2D human pose estimation method and PoseLifter in a cascade fashion, as shown in Figure~\ref{fig4}. Similar to other 3D human pose estimation studies, we assume that a single RGB image and a bounding box that contains the target subject in the image are provided as inputs. First, we crop the input image using the bounding box and then resize the resulting image to $256\times256$. The resized image is then fed into a state-of-the-art 2D human pose estimator to obtain a 2D pose. Thereafter, the obtained 2D pose is transformed into the original image's coordinate system and fed into our PoseLifter to yield the canonical root depth $\tilde{R}_{z}$ and the root-relative pose $\hat{\mathbf{P}}$. If focal length information is available, the canonical root depth can be converted to the absolute root coordinates $\mathbf{R}=(R_{x},R_{y},R_{z})$ through Equations~(\ref{eq5}) and~(\ref{eq6}). Finally, the absolute root coordinates and the root-relative pose are transformed into an absolute 3D pose using $\mathbf{P}_{i}=\mathbf{R}+\hat{\mathbf{P}}_{i}$.

\subsection{2D human pose estimation}
\label{sec4.1}

The recently proposed heatmap regression network~\cite{Xiao2018} and integral regression~\cite{Sun2018} are used for 2D human pose estimation from a single RGB image. The method presented in~\cite{Xiao2018} removes the last two layers of a residual neural network (\textit{i.e.}, ResNet)~\cite{He2016} and adds three deconvolution layers and a $1\times{}1$ convolution layer to the back. This modified network receives an image with a size of $3\times256\times256$ and generates output heatmaps with a size of $J\times64\times64$, which intuitively represent probability distributions for the 2D locations of each joint.

The process of obtaining 2D joint coordinates from heatmaps usually depends on the argmax operation. Argmax has two drawbacks: (1) vulnerability to quantization errors and (2) nondifferentiability. Therefore, a method that can directly calculate the 2D joint coordinates from a heatmap with sub-pixel accuracy in a differentiable manner has been proposed for integral regression~\cite{Sun2018}. The basic concept of this method is to normalize a heatmap to a probability distribution and then apply the expectation operation to the result.

In this study, we construct a 2D human pose estimator by attaching an integral module to the back of the heatmap regression network based on the ResNet backbone. This estimator outputs the heatmaps and 2D joint coordinates from an input RGB image. The mean squared error (MSE) and L1 losses are used to supervise the heatmaps and 2D coordinates, respectively, as a cost function for estimator learning.

\subsection{Synthesizing the input for training PoseLifter}
\label{sec4.2}

The result of the 2D pose estimator is used as the input for our PoseLifter to generate an absolute 3D pose. Such input 2D pose is imperfect and includes an estimation error. We propose to model the estimation error and use the sampled realistic 2D pose based on the error distribution for the learning of PoseLifter to make the system robust to error.

\begin{figure}
\centering
\includegraphics[width=0.95\linewidth]{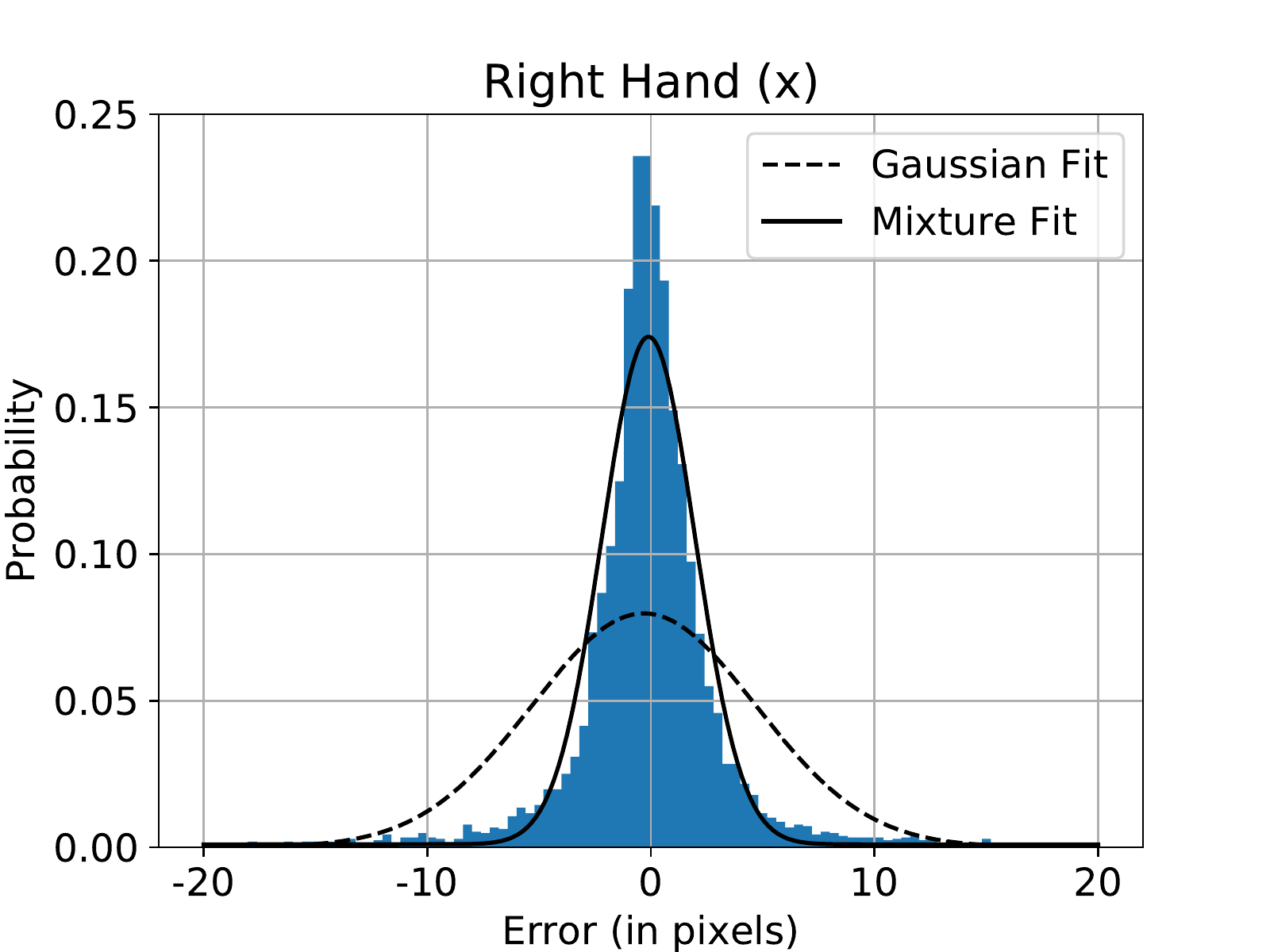}
\caption{The error distribution of the \textit{right hand} joint in the $x$ axis is shown in the blue histogram. The dotted and solid lines show the single Gaussian model and the proposed mixture model fitted to such empirical data, respectively.}
\label{fig5}
\end{figure}

To obtain the error statistics of 2D pose estimation, we learn a 2D human pose estimator by using data that is generated by excluding one of the human subjects that constitute the training set. The estimation error is obtained by applying the learned model to the excluded human subject data, as shown in Figure~\ref{fig5}. We propose that a simple mixture model consisting of Gaussian and uniform distributions is well suited for such an empirical error distribution. Following the error taxonomy in~\cite{Ronchi2017}, the Gaussian and uniform distributions can account for the error of the inlier joint, such as \textit{good} or \textit{jitter}, and the error of the outlier joint, such as \textit{inversion} or \textit{miss}, respectively.

The mixture model for error $\mathbf{e}=(e_x,e_y)$ is as follows:
\begin{equation}\label{eq8}
p(\mathbf{e})=\gamma\frac{\exp\left(-\frac{1}{2}(\mathbf{e}-\bm{\mu})^{T}\bm{\Sigma}^{-1}(\mathbf{e}-\bm{\mu})\right)}{2\pi\sqrt{|\bm{\Sigma}|}}+(1-\gamma)\frac{1}{v},
\end{equation}
where $\gamma$ is the mixing parameter, $v$ is the normalization constant of the uniform distribution, and $\bm{\mu}=(\mu_x,\mu_y)$ and $\bm{\Sigma}=diag(\sigma_x^2,\sigma_y^2)$ represent the mean vector and covariance matrix of the Gaussian distribution, respectively. We initially set $v$ to $100\times100=10000$, assuming that the pixel range of the error due to the outlier is $[-50,50]$. The remaining parameters of the mixture model are determined by minimizing the negative log likelihood as follows:
\begin{equation}\label{eq9}
NLL=-\sum_{i}\log{}p(\mathbf{e}^{(i)}),
\end{equation}
where superscript $i$ is the index of the sample. This minimization process can be realized by using the expectation maximization (EM) algorithm~\cite{Dempster1977}.

To obtain the initial estimate of the mean and covariance of the Gaussian distribution, a single Gaussian is fitted to the given error data, and $\gamma$ is initialized to 0.9. Figure~\ref{fig5} shows that our mixture model effectively explains the empirical error data. The error model obtained through the preceding procedure is used to synthesize the 2D input pose of the training set required for PoseLifter learning, thereby making the resulting PoseLifter robust to real 2D pose estimation errors.

\section{Experimental results}
\label{sec5}

\subsection{Implementation details}
\label{sec5.1}

PyTorch~\cite{Paszke2017} is used as the deep learning framework in all our experiments. Our code to reproduce all the results of this paper is available in~\footnote{\url{https://github.com/juyongchang/PoseLifter}}. PoseLifter and the 2D pose estimator are learned separately through the following processes.

\textbf{PoseLifter.} First, the cost function of Equation~(\ref{eq7}) is minimized using the Rmsprop optimization algorithm \cite{Tieleman2012} for PoseLifter learning. Learning rate, batch size, and number of epochs are set to $10^{-3}$, 64, and 300, respectively. Learning rate is reduced to $10^{-4}$ after 200 epochs. Except for random horizontal flipping, no data augmentation is performed in 2D-to-3D pose lifting experiments. In the case of 3D human pose estimation from a single RGB image, the input synthesis process in Section~\ref{sec4.2} functions as an additional data augmentation procedure.

\textbf{2D pose estimator.} The ResNet152 backbone-based network for 2D pose estimation is initialized with pre-trained weights from the ImageNet dataset~\cite{Russakovsky2015}. Next, the sum of the MSE heatmap and L1 coordinate losses in Section~\ref{sec4.1} is minimized using the Rmsprop algorithm. In this case, learning rate, batch size, and number of epochs are set to $10^{-4}$, 48, and 60, respectively. Learning rate is reduced to $10^{-5}$ after 30 epochs. For data augmentation, a small random translation of $[-4,4]$ pixel range, random horizontal flipping, and 40\% random color jittering are applied to the input RGB image.

\subsection{Dataset and evaluation metrics}
\label{sec5.2}

\begin{table}[]
\small
\centering
\setlength\tabcolsep{1.0pt}
\def\arraystretch{1.1}
\begin{tabular}{L{2.3cm}C{1.5cm}C{2.0cm}C{2.1cm}}
\specialrule{.1em}{.05em}{.05em}
\multicolumn{3}{c}{Dataset}                                            & Focal length       \\
\specialrule{.1em}{.05em}{.05em}
Human3.6M                     & \multicolumn{2}{c}{Training/Test sets} & 1146.79$\pm{}$2.28 \\ \hline
\multirow{4}{*}{MPI-INF-3DHP} & \multicolumn{2}{c}{Training set}       & 1497.39$\pm{}$2.95 \\
                              & \multirow{3}{*}{Test set} & StudioGS   & 1499.21            \\
                              &                           & StudioNoGS & 1499.21            \\
                              &                           & Outdoor    & 1683.98            \\
\specialrule{.1em}{.05em}{.05em}
\end{tabular}
\caption{The focal length ($\alpha$) values of the cameras used to obtain the images of the two datasets are shown. The unit of focal length is pixels}
\label{table1}
\end{table}

The datasets and metrics used to evaluate the performance of the proposed method are as follows.

\textbf{Human3.6M.} The Human3.6M dataset~\cite{Ionescu2014} is used as the primary dataset for evaluating the proposed method. This dataset consists of approximately 3.6M RGB images and their corresponding 2D and 3D poses. A motion capture system that includes 4 cameras is used to obtain the poses of 11 actors performing 15 activities. In addition, the MPII dataset~\cite{Andriluka2014} is used for the learning of the 2D pose estimator. This dataset consists of approximately 25K in-the-wild images and their corresponding 2D pose information.

For the evaluation metric, we use the mean per joint position error (MPJPE), which is defined as the mean of the Euclidean distances between the corresponding joints after aligning the root joints of ground truth and the estimated 3D pose. Moreover, the PA-MPJPE, which is calculated after applying Procrustes alignment~\cite{Gower1975} to the two 3D poses, is adopted as an additional metric. The two metrics are used to evaluate the root-relative pose computed using our method. To evaluate the absolute location of the root joint, we introduce the mean of the Euclidean distance between the prediction $\mathbf{R}$ and the ground truth $\mathbf{R}^{*}$, \textit{i.e.}, the mean of the root position error (MRPE), as a new metric:
\begin{equation}\label{eq10}
MRPE=\frac{1}{N}\sum_{i=1}^{N}||\mathbf{R}^{(i)}-\mathbf{R}^{(i)*}||_{2},
\end{equation}
where superscript $i$ is the index of the sample, and $N$ denotes the total number of test samples.

Two protocols adopted from existing works are used to evaluate the proposed method. In \textit{Protocol 1}, Subjects 1, 5, 6, 7, and 8 and Subjects 9 and 11 are used as the training and test sets, respectively. Evaluation is performed through the MPJPE metric. In \textit{Protocol 2}, Subjects 1, 5, 6, 7, 8, and 9 and Subject 11 are adopted as the training and test sets, respectively. The PA-MPJPE metric is used for evaluation.

\textbf{MPI-INF-3DHP.} We additionally use the MPI-INF-3DHP dataset~\cite{Mehta2017} to evaluate the proposed method. This dataset consists of approximately 1.3M frames acquired using a commercial marker-less motion capture system with multiple cameras. Approximately 190K frames obtained by sampling the training set every 5 frames are used for the learning of the proposed model. Meanwhile, the original test set that consists of 2935 frames is used for evaluation.

\begin{figure}
\centering
\includegraphics[width=0.95\linewidth]{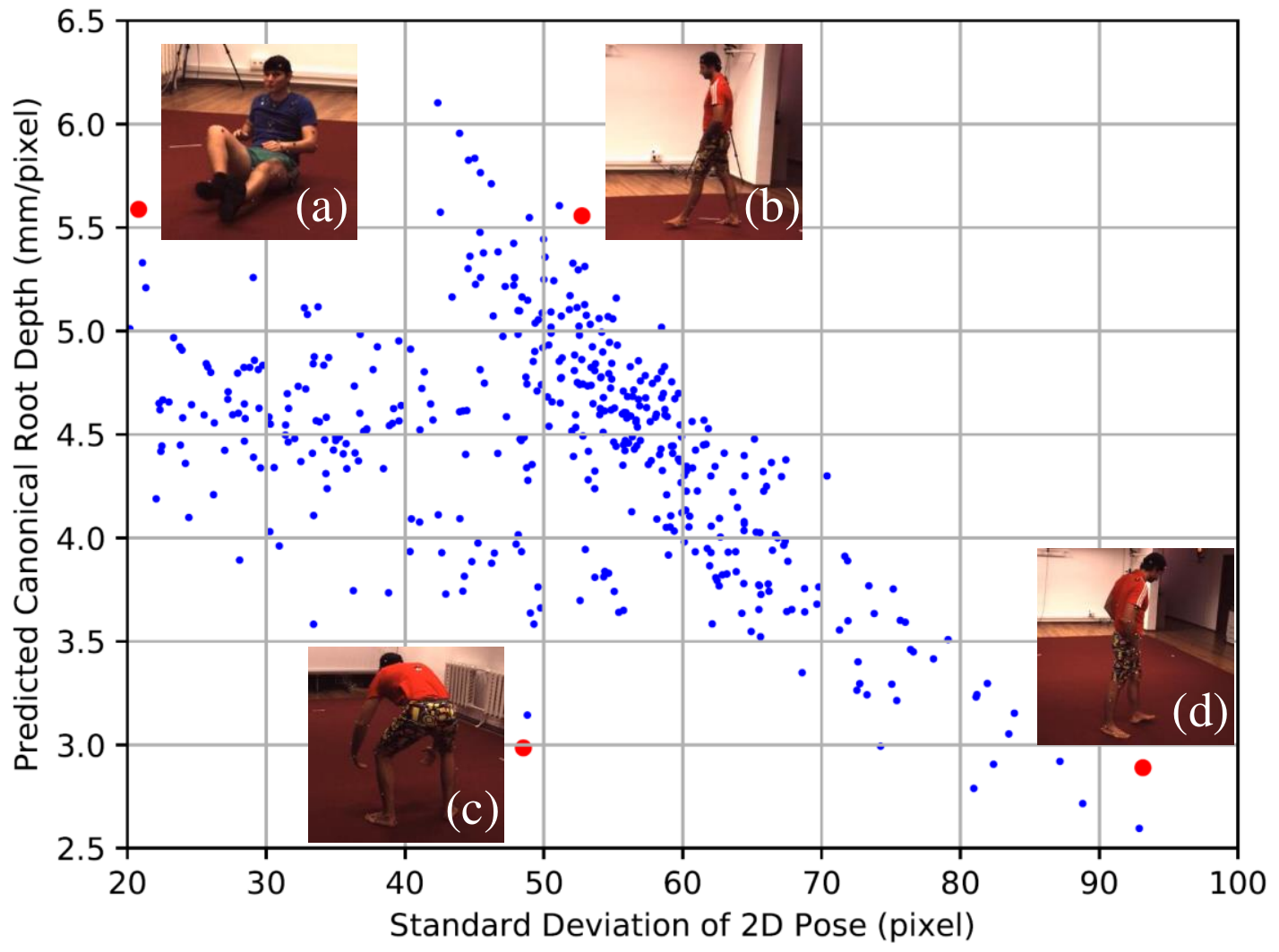}
\caption{The standard deviation ($\sigma$) of the 2D pose of the test samples in the Human3.6M dataset and the canonical root depth ($\tilde{R}_{z}$) predicted by our PoseLifter are plotted.}
\label{fig6}
\end{figure}

In addition, 3DPCK, which extends the existing percentage of correct keypoints (PCK)~\cite{Tompson2014,Toshev2014} to 3D, and the area under curve (AUC), which is calculated for several PCK thresholds, are used as evaluation metrics. To compute 3DPCK, whether the distance between the corresponding joints is less than $150 mm$ is checked after the roots of the predicted and ground truth 3D poses are aligned. The data acquisition environment for the test set can be divided into a studio with a green screen (StudioGS), a studio without a green screen (StudioNoGS), and outdoor (Outdoor). 3DPCK and AUC results are reported for each case.

Table~\ref{table1} shows the focal length information of the cameras used to acquire the Human3.6M and MPI-INF-3DHP datasets. Unlike Human3.6M datasets with similar focal lengths, the MPI-INF-3DHP dataset has considerably different focal length values. Specifically, in the case of the Outdoor data, its focal length value is 1683.98 pixels, which is approximately 11\% different from those of other test data (1499.21 pixels) or training data (1497.39 pixels).

\subsection{2D-to-3D pose lifting}
\label{sec5.3}

\begin{table}
\small
\centering
\setlength\tabcolsep{1.0pt}
\def\arraystretch{1.1}
\begin{tabular}{L{3.1cm}C{1.6cm}C{1.6cm}C{1.6cm}}
\specialrule{.1em}{.05em}{.05em}
Method & MPJPE & PA-MPJPE & MRPE \\
\specialrule{.1em}{.05em}{.05em}
Baseline & 44.86 & 30.38 & 510.88 \\
+ Location & 39.04 & 30.25 & 368.89 \\
+ Scale & 44.29 & 30.04 & 113.38 \\
+ Location \& scale & \textbf{38.38} & \textbf{29.91} & \textbf{98.84} \\
\hline
Ramakrishna~\cite{Ramakrishna2012} & - & 89.50 & - \\
Dai~\cite{Dai2014} & - & 72.98 & - \\
Zhou~\cite{Zhou2015} & - & 50.04 & - \\
Zhou~\cite{Zhou2016} & - & 49.64 & - \\
Moreno-Noguer\cite{Moreno-Noguer2017} & - & 62.17 & - \\
Martinez~\cite{Martinez2017} & - & 37.10 & - \\
\specialrule{.1em}{.05em}{.05em}
\end{tabular}
\caption{2D-to-3D pose lifting performances are shown for our PoseLifter and its variants and other existing methods, in which the Human3.6M dataset is used. Protocol 1 is adopted and the unit of all numbers is $mm$.}
\label{table2}
\end{table}

In this subsection, we present the performance of the proposed PoseLifter for 2D-to-3D pose lifting.

\textbf{Canonical root depth estimation.} Under the perspective projection assumption, the canonical root depth is inversely proportional to the image scale for human subjects with a constant real scale. This inverse proportion property is learned implicitly by the proposed PoseLifter, as shown in Figure~\ref{fig6}, where the predicted canonical root depth is approximately in inverse proportion to the human scale in image space (\textit{i.e.}, standard deviation of the 2D pose). However, the various postures of the human subject lead to variations in real scale, making the estimation of the canonical root depth using only the inverse proportion property difficult. This problem can be resolved by our method, which allows us to implicitly compute the variation of the real scale using the input normalized 2D pose for a more accurate estimation of the canonical root depth. For example, for Figures~\ref{fig6}(a) and (b) with different image scales, the proposed method estimates similar canonical root depths, taking into account variations in real scales.

\textbf{Performance analysis.} We first evaluate the performance of our PoseLifter for 2D-to-3D pose lifting. To achieve this, we train and test our PoseLifter using ground truth 2D pose data. Table~\ref{table2} presents the performance of our PoseLifter with and without location and scale information. To implement the latter, we modify our normalization layer to output only normalized 2D pose $\{\hat{\mathbf{p}}_{i}\}_{i=1}^{J}$, except for mean $\mathbf{u}$ and standard deviation $\sigma$. The experimental results indicate that using location and scale information is essential in estimating an accurate absolute root position. In this case, MRPE is $98.84 mm$. Moreover, the location and scale information required for absolute root estimation considerably reduces the root-relative pose error (\textit{i.e.}, MPJPE) from $44.86 mm$ to $38.38 mm$. This result demonstrates the effectiveness of our proposed PoseLifter in estimating the absolute pose (\textit{i.e.}, absolute root coordinates and root-relative pose).

\textbf{Quantitative comparison.} Subsequently, a quantitative comparison is performed between the proposed method and existing methods for 2D-to-3D pose lifting. The results are provided in Table~\ref{table2}. The proposed method outperforms optimization-based methods~\cite{Dai2014,Ramakrishna2012,Zhou2015,Zhou2016} and the more recent regression-based methods~\cite{Martinez2017,Moreno-Noguer2017} in terms of root-relative 3D pose estimation while allowing the acquisition of absolute location information, which the other methods are incapable of doing.

\begin{table}
\small
\centering
\setlength\tabcolsep{1.0pt}
\def\arraystretch{1.1}
\begin{tabular}{L{2.1cm}C{1.6cm}C{1.3cm}C{1.6cm}C{1.3cm}}
\specialrule{.1em}{.05em}{.05em}
Input 2D pose & Loc.\&Scale & MPJPE & PA-MPJPE & MRPE \\
\specialrule{.1em}{.05em}{.05em}
GT & & 64.97 & 45.47 & 680.38 \\
GT & $\checkmark$ & 61.53 & 45.45 & 239.35 \\
2D estimate & $\checkmark$ & 58.90 & 43.54 & 209.44 \\
Single Gaussian & $\checkmark$ & 56.01 & 44.78 & 164.51 \\
Mixture model & $\checkmark$ & \textbf{53.14} & \textbf{42.63} & \textbf{144.24} \\
\specialrule{.1em}{.05em}{.05em}
\end{tabular}
\caption{The results of our cascade approach are presented along the input generating strategy for PoseLifter learning. The Human3.6M dataset is used. ``Loc.\&Scale'' indicates that the 2D location and scale information is utilized in PoseLifter. ``2D estimate'' means to use the output of the 2D pose estimator as a training set. ``Single Gaussian'' and ``Mixture model'' represent the error model used for synthesizing the 2D pose.}
\label{table3}
\end{table}

\subsection{Cascading with 2D pose estimator}
\label{sec5.4}

In this subsection, we present the performance of the proposed cascade approach with a 2D pose estimator in estimating 3D human pose from a single RGB image. Table~\ref{table3} provides the results for the Human3.6M dataset. As with 2D-to-3D pose lifting, the approximate 2D location and scale information of the target subject boosts the performance of the root-relative pose and the absolute root location, which are evaluated through MPJPE and MRPE, respectively.

\textbf{Synthesizing 2D pose for PoseLifter.} As shown in Table~\ref{table3}, the input 2D pose data used for PoseLifter learning play an important role in the performance of 3D pose estimation. First, the use of ground truth 2D pose provides the worst results. This result is improved by using real 2D pose estimates obtained by applying the 2D pose estimator to training images for PoseLifter learning. Evidently, the use of realistic input data for learning improves the performance of the model. However, such input data are fixed for a given specific training data, thereby limiting their variability. One approach to overcome this problem is to synthesize the input data in accordance with the underlying distribution. The experiment that uses the error model obtained by analyzing the actual 2D pose error statistics demonstrates that the single Gaussian model does not substantially improve performance. Figure~\ref{fig5} shows that the single Gaussian model estimates a larger standard deviation than necessary because of outliers. By contrast, our proposed mixture model yields considerably improved results for all the evaluation metrics. The results clearly prove our hypothesis that the synthesis of realistic inputs is beneficial for the performance of the model.

\begin{table}
\small
\centering
\setlength\tabcolsep{1.0pt}
\def\arraystretch{1.1}
\begin{tabular}{L{1.9cm}C{1.5cm}C{1.5cm}C{1.5cm}C{1.5cm}}
\specialrule{.1em}{.05em}{.05em}
Method & MRPE & MRPE-X & MRPE-Y & MRPE-Z \\
\specialrule{.1em}{.05em}{.05em}
Mehta~\cite{Mehta2017} & 164.0 & \textbf{20.4} & \textbf{20.2} & 157.5 \\
Moon~\cite{Moon2019b} & \textbf{120.0} & 23.3 & 23.0 & \textbf{108.1} \\
Ours & 144.2 & \textbf{20.4} & 21.0 & 137.0 \\
\specialrule{.1em}{.05em}{.05em}
\end{tabular}
\caption{Quantitative results of root location estimation are given for the methods in~\cite{Mehta2017,Moon2019b} and ours, in which the Human3.6M dataset is used. MRPE-X, MRPE-Y, and MRPE-Z represent the mean of the errors in the $X$, $Y$, and $Z$ axes, respectively.}
\label{table4}
\end{table}

\begin{table*}
\scriptsize
\centering
\setlength\tabcolsep{1.0pt}
\def\arraystretch{1.1}
\begin{tabular}{L{3.1cm}C{0.79cm}C{0.79cm}C{0.79cm}C{0.79cm}C{0.79cm}C{0.79cm}C{0.79cm}C{0.79cm}C{0.79cm}C{0.79cm}C{0.79cm}C{0.79cm}C{0.79cm}C{0.79cm}C{0.79cm}C{0.79cm}C{0.4cm}}
\specialrule{.1em}{.05em}{.05em}
\textbf{Protocol 1} & Direct. & Discuss & Eating & Greet & Phone & Pose & Purch. & Sit & SitD. & Smoke & Photo & Wait & Walk & WalkD. & WalkT. & Avg. & GT \\
\specialrule{.1em}{.05em}{.05em}
\multicolumn{18}{l}{\textbf{\textit{Direct approaches}}} \\
Kanazawa CVPR'18~\cite{Kanazawa2018} & - & - & - & - & - & - & - & - & - & - & - & - & - & - & - & 88.0 & \\
Mehta 3DV'17~\cite{Mehta2017} & 57.5 & 68.6 & 59.6 & 67.3 & 78.1 & 56.9 & 69.1 & 98.0 & 117.5 & 69.5 & 82.4 & 68.0 & 55.3 & 76.5 & 61.4 & 72.9 & \\
Pavlakos CVPR'17~\cite{Pavlakos2017} & 67.4 & 72.0 & 66.7 & 69.1 & 72.0 & 65.0 & 68.3 & 83.7 & 96.5 & 71.7 & 77.0 & 65.8 & 59.1 & 74.9 & 63.2 & 71.9 & \\
S\'{a}r\'{a}ndi ECCVW'18~\cite{Sarandi2018} & 63.6 & 65.5 & 56.0 & 62.1 & 64.0 & 60.7 & 64.8 & 76.7 & 93.0 & 63.3 & 69.7 & 62.0 & 68.8 & 61.3 & 54.1 & 65.7 & \\
Zhou ICCV'17~\cite{Zhou2017} & 54.8 & 60.7 & 58.2 & 71.4 & 62.0 & 53.8 & 55.6 & 75.2 & 111.6 & 64.1 & 65.5 & 66.0 & 63.2 & 51.4 & 55.3 & 64.9 & $\checkmark$ \\
Sun ICCV'17~\cite{Sun2017} & 52.8 & 54.8 & 54.2 & 54.3 & 61.8 & 53.1 & 53.6 & 71.7 & 86.7 & 61.5 & 67.2 & 53.4 & 47.1 & 61.6 & 53.4 & 59.1 & \\
Yang CVPR'18~\cite{Yang2018} & 51.5 & 58.9 & 50.4 & 57.1 & 62.1 & 49.8 & 52.7 & 69.2 & 85.2 & 57.4 & 65.4 & 58.4 & 60.1 & \textbf{43.6} & 47.7 & 58.6 & $\checkmark$ \\
Moon ICCV'19~\cite{Moon2019b} & 51.5 & 56.8 & 51.2 & 52.2 & 55.2 & 47.7 & 50.9 & 63.3 & 69.9 & 54.2 & 57.4 & 50.4 & 42.5 & 57.5 & 47.7 & 54.4 & \\
S\'{a}r\'{a}ndi ECCVW'18~\cite{Sarandi2018}* & 49.1 & 54.6 & 50.4 & 50.7 & 54.8 & 47.4 & 50.1 & 67.5 & 78.4 & 53.1 & 57.4 & 50.7 & 54.0 & 46.1 & \textbf{40.1} & 54.2 & \\
Sun ECCV'18~\cite{Sun2018} & \textbf{47.5} & \textbf{47.7} & \textbf{49.5} & \textbf{50.2} & \textbf{51.4} & \textbf{43.8} & \textbf{46.4} & \textbf{58.9} & \textbf{65.7} & \textbf{49.4} & \textbf{55.8} & \textbf{47.8} & \textbf{38.9} & 49.0 & 43.8 & \textbf{49.6} & $\checkmark$ \\
\hline
\multicolumn{18}{l}{\textbf{\textit{Cascade approaches}}} \\
Zhou TPAMI'18~\cite{Zhou2018} & 68.7 & 74.8 & 67.8 & 76.4 & 76.3 & 84.0 & 70.2 & 88.0 & 113.8 & 78.0 & 98.4 & 90.1 & 62.6 & 75.1 & 73.6 & 79.9 & \\
Martinez ICCV'17~\cite{Martinez2017} & 51.8 & 56.2 & 58.1 & 59.0 & 69.5 & 55.2 & 58.1 & 74.0 & 94.6 & 62.3 & 78.4 & 59.1 & 49.5 & 65.1 & 52.4 & 62.9 & \\
Fang AAAI'18~\cite{Fang2018} & 50.1 & 54.3 & 57.0 & 57.1 & 66.6 & 53.4 & 55.7 & 72.8 & 88.6 & 60.3 & 73.3 & 57.7 & 47.5 & 62.7 & 50.6 & 60.4 & \\
Ours & \textbf{44.8} & \textbf{48.2} & \textbf{48.5} & \textbf{51.5} & \textbf{54.5} & \textbf{47.9} & \textbf{47.8} & \textbf{60.7} & \textbf{76.4} & \textbf{52.5} & \textbf{64.4} & \textbf{50.8} & \textbf{39.0} & \textbf{55.3} & \textbf{42.2} & \textbf{52.5} & \\
\specialrule{.1em}{.05em}{.05em}
\textbf{Protocol 2} & Direct. & Discuss & Eating & Greet & Phone & Pose & Purch. & Sit & SitD. & Smoke & Photo & Wait & Walk & WalkD. & WalkT. & Avg. & GT \\
\specialrule{.1em}{.05em}{.05em}
\multicolumn{17}{l}{\textbf{\textit{Direct approaches}}} \\
Kanazawa CVPR'18~\cite{Kanazawa2018} & - & - & - & - & - & - & - & - & - & - & - & - & - & - & - & 56.8 & \\
Sun ECCV'18~\cite{Sun2018} & 36.9 & 36.2 & 40.6 & 40.4 & 41.9 & 34.9 & 35.7 & 50.1 & 59.4 & 40.4 & 44.9 & 39.0 & 30.8 & 39.8 & 36.7 & 40.6 & $\checkmark$ \\
Yang CVPR'18~\cite{Yang2018} & \textbf{26.9} & \textbf{30.9} & \textbf{36.3} & 39.9 & 43.9 & \textbf{28.8} & \textbf{29.4} & 36.9 & 58.4 & 41.5 & 47.4 & \textbf{30.5} & 42.5 & \textbf{29.5} & 32.2 & 37.7 & $\checkmark$ \\
Moon ICCV'19~\cite{Moon2019b} & 32.5 & 31.5 & 41.5 & \textbf{36.7} & \textbf{36.3} & 31.9 & 33.2 & \textbf{36.5} & \textbf{44.4} & \textbf{36.7} & \textbf{38.7} & 31.2 & \textbf{25.6} & 37.1 & \textbf{30.5} & \textbf{35.2} & \\
\hline
\multicolumn{17}{l}{\textbf{\textit{Cascade approaches}}} \\
Ramakrishna ECCV'12~\cite{Ramakrishna2012} & 137.4 & 149.3 & 141.6 & 154.3 & 157.7 & 141.8 & 158.1 & 168.6 & 175.6 & 160.4 & 158.9 & 161.7 & 174.8 & 150.0 & 150.2 & 157.3 & \\
Bogo ECCV'16~\cite{Bogo2016} & 62.0 & 60.2 & 67.8 & 76.5 & 92.1 & 73.0 & 75.3 & 100.3 & 137.3 & 83.4 & 77.0 & 77.3 & 79.7 & 86.8 & 87.7 & 82.3 & \\
Moreno-Noguer CVPR'17~\cite{Moreno-Noguer2017} & 66.1 & 61.7 & 84.5 & 73.7 & 65.2 & 60.9 & 67.3 & 103.5 & 74.6 & 92.6 & 67.2 & 69.6 & 78.0 & 71.5 & 73.2 & 74.0 & \\
Zhou TPAMI'18~\cite{Zhou2018} & 47.9 & 48.8 & 52.7 & 55.0 & 56.8 & 49.0 & 45.5 & 60.8 & 81.1 & 53.7 & 65.5 & 51.6 & 50.4 & 54.8 & 55.9 & 55.3 & \\
Martinez ICCV'17~\cite{Martinez2017} & 39.5 & 43.2 & 46.4 & 47.0 & 51.0 & 41.4 & 40.6 & 56.5 & 69.4 & 49.2 & 56.0 & 45.0 & 38.0 & 49.5 & 43.1 & 47.7 & \\
Fang AAAI'18~\cite{Fang2018} & 38.2 & 41.7 & 43.7 & 44.9 & 48.5 & 40.2 & 38.2 & 54.5 & 64.4 & 47.2 & 55.3 & 44.3 & 36.7 & 47.3 & 41.7 & 45.7 & \\
Ours & \textbf{32.1} & \textbf{34.9} & \textbf{43.4} & \textbf{36.9} & \textbf{35.4} & \textbf{35.1} & \textbf{30.8} & \textbf{34.3} & \textbf{57.3} & \textbf{40.4} & \textbf{44.9} & \textbf{35.1} & \textbf{24.9} & \textbf{46.6} & \textbf{30.0} & \textbf{37.7} & \\
\specialrule{.1em}{.05em}{.05em}
\end{tabular}
\caption{A quantitative comparison of our approach and other recent methods for 3D human pose estimation from a single RGB image is illustrated. ``*'' indicates additional Pascal VOC dataset~\cite{Everingham2010} is used for training. ``GT'' means that the root's ground truth depth has been used during the estimation process. The Human3.6M dataset is used. MPJPE and PA-MPJPE are adopted for Protocols 1 and 2, respectively.}
\label{table5}
\end{table*}

\textbf{Root location estimation.} To estimate the absolute location of the root, our learning-based method relies on a large-scale 2D/3D pose dataset. In~\cite{Mehta2017}, a method was proposed for analytically calculating the root position from given 2D and root-relative 3D poses without learning. This method assumes a weak perspective projection and is based on a linear least square formulation. We refer to the supplementary material of~\cite{Mehta2017} for the formula for calculating the depth $Z$ of the root. The remaining $X$ and $Y$ coordinates are obtained using a back-projection formula. Table~\ref{table4} presents a quantitative comparison between the analytic approach presented in~\cite{Mehta2017} and our learning-based approach, in which 2D and 3D root-relative poses obtained through our method are used for fair comparison. Our method exhibits a better $Z$ error than the analytic method. For the errors in the $X$ and $Y$ directions, our method produces nearly the same result as the analytic method, which relies on the constraint that the 3D point should be located on the back-projecting ray. This finding shows that our pose-lifting method implicitly enforces such constraints through learning. The recently proposed learning-based method in~\cite{Moon2019b} shows better results than ours. We believe this is because our method only relies on 2D poses, while the other approach can utilize image features.

\begin{figure*}[t]
\centering
\includegraphics[width=.46\linewidth]{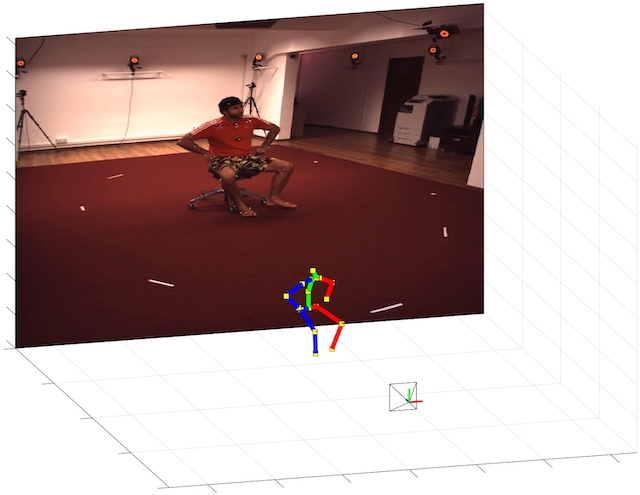}
\includegraphics[width=.46\linewidth]{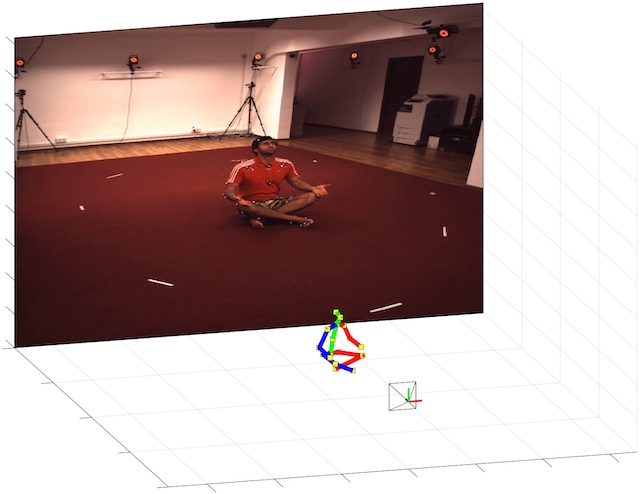}
\includegraphics[width=.46\linewidth]{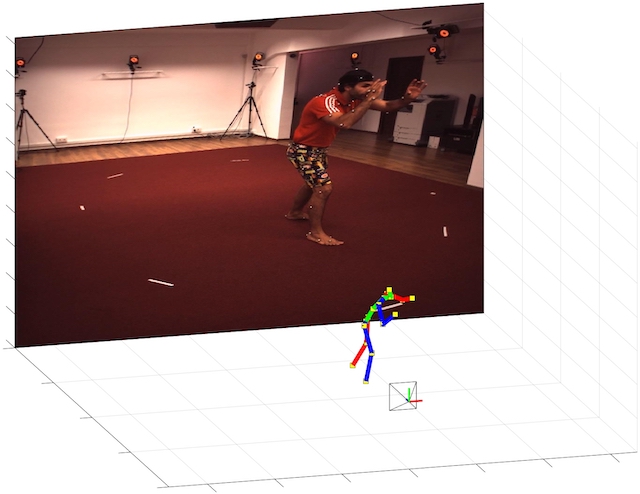}
\includegraphics[width=.46\linewidth]{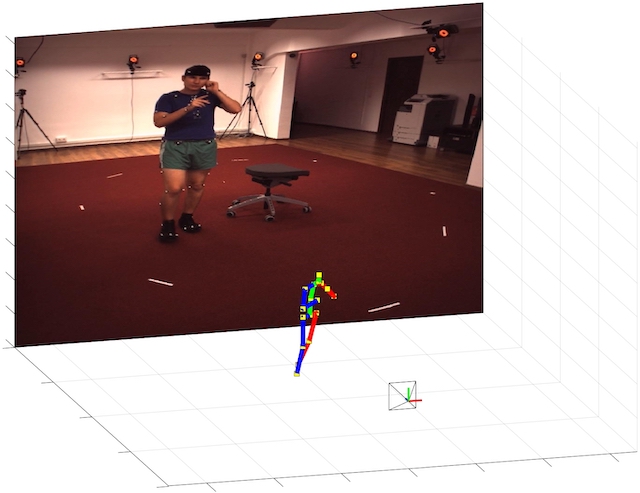}
\includegraphics[width=.46\linewidth]{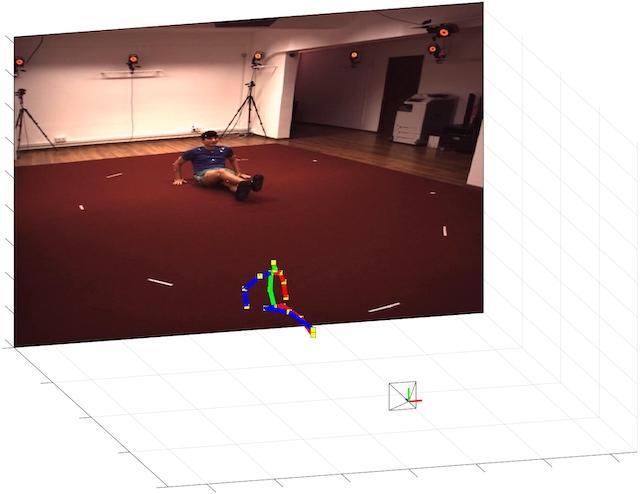}
\includegraphics[width=.46\linewidth]{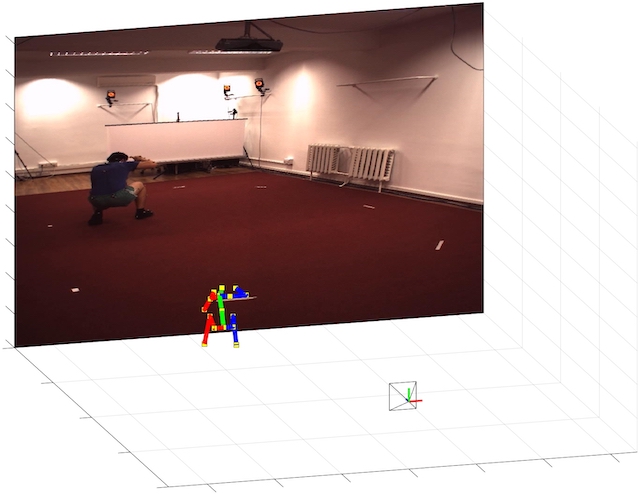}
\caption{Qualitative results of our method are shown for the Human3.6M dataset.}
\label{fig7}
\end{figure*}

\textbf{Quantitative comparison.} Table~\ref{table5} provides the quantitative results of the performance of recently proposed methods for 3D human pose estimation from a single RGB image. The proposed method achieves comparable performance with state-of-the-art methods for Protocols 1 and 2. In particular, our method outperforms all cascade approaches~\cite{Bogo2016,Fang2018,Martinez2017,Moreno-Noguer2017,Ramakrishna2012} that consist of a sequential combination of a 2D pose estimator and a 3D lifter. Two direct methods in~\cite{Moon2019b,Sun2018} yield better results than ours for Protocols 1 and 2, respectively. However, several methods~\cite{Sun2018,Yang2018,Zhou2017}, including the aforementioned method~\cite{Sun2018}, require the ground truth for the absolute depth of the root joint to produce a 3D pose result. This constraint is attributed to these methods using a back-projection formula to transform the joint's estimated image coordinates $x$ and $y$ to 3D coordinates $X$ and $Y$, which require the absolute depth of each joint. By contrast, our method exhibits the advantage of not requiring the ground truth because the absolute location of the root is estimated. The qualitative results for some of the images in the test dataset are shown in Figure~\ref{fig7}.

\begin{table*}
\scriptsize
\centering
\setlength\tabcolsep{1.0pt}
\def\arraystretch{1.1}
\begin{tabular}{L{2.6cm}C{1.5cm}C{1.0cm}C{1.5cm}C{1.0cm}C{1.5cm}C{1.0cm}C{1.0cm}}
\specialrule{.1em}{.05em}{.05em}
Method & 3D Dataset & StudioGS & StudioNoGS & Outdoor & All 3DPCK & AUC & MRPE \\
\specialrule{.1em}{.05em}{.05em}
Yang~\cite{Yang2018} & H3.6M & - & - & - & 69.0 & 32.0 & - \\
Zhou~\cite{Zhou2017} & H3.6M & 71.1 & 64.7 & \textbf{72.7} & 69.2 & 32.5 & - \\
Ours w/o canonical depth & H3.6M & 81.1& 73.5 & 72.5 & 76.2 & 40.0 & 920.1 \\
Ours & H3.6M & \textbf{81.6} & \textbf{73.6} & 72.5 & \textbf{76.5} & \textbf{40.2} & \textbf{421.3} \\
\hline
Mehta~\cite{Mehta2017} & INF & 84.1 & 68.9 & 59.6 & 72.5 & 36.9 & - \\
Mehta~\cite{Mehta2017} & INF+H3.6M & 84.6 & 72.4 & 69.7 & 76.5 & 40.8 & - \\
Mehta~\cite{Mehta2017b} & INF+H3.6M & - & - & - & 76.6 & 40.4 & - \\
Kanazawa~\cite{Kanazawa2018} & INF+H3.6M & - & - & - & 72.9 & 36.5 & - \\
Ours w/o canonical depth & INF & 91.3 & 78.2 & 64.6 & 79.9 & 42.3 & 296.9 \\
Ours & INF & \textbf{91.4} & \textbf{82.9} & \textbf{73.7} & \textbf{83.9} & \textbf{45.0} & \textbf{217.4} \\
\specialrule{.1em}{.05em}{.05em}
\end{tabular}
\caption{The comparison results of the proposed and other methods are shown for the MPI-INF-3DHP dataset. ``Ours w/o canonical depth'' means that the canonical depth representation is not used.}
\label{table6}
\end{table*}

\textbf{MPI-INF-3DHP.} Subsequently, we present the evaluation results of the proposed method for the MPI-INF-3DHP dataset, as shown in Table~\ref{table6}. We use an additional 2D pose dataset, \textit{i.e.}, MPII, to learn the 2D pose estimator following~\cite{Mehta2017}. PoseLifter is learned using either Human3.6M or MPI-INF-3DHP. The input synthesis method presented in Section~\ref{sec4.2} is applied to this process.

In contrast with the Human3.6M dataset, the focal length parameters of the cameras used to acquire the training and test sets for the MPI-INF-3DHP dataset are considerably different, which is shown in Table~\ref{table1}. By the focal length ambiguity, this allows a single input 2D pose to be mapped to multiple absolute root depths corresponding to different focal length parameters. Therefore, learning to regress the absolute root depth directly, based on a dataset containing different focal length images, becomes a seriously ill-posed problem. The \textit{canonical depth representation} proposed in Section~\ref{sec3.2} resolves this problem by causing our network to regress the canonical root depth normalized by the focal length, as shown in Table~\ref{table6}. The use of the canonical depth representation significantly reduces MRPE.

\begin{figure*}[t]
\centering
\includegraphics[width=.46\linewidth]{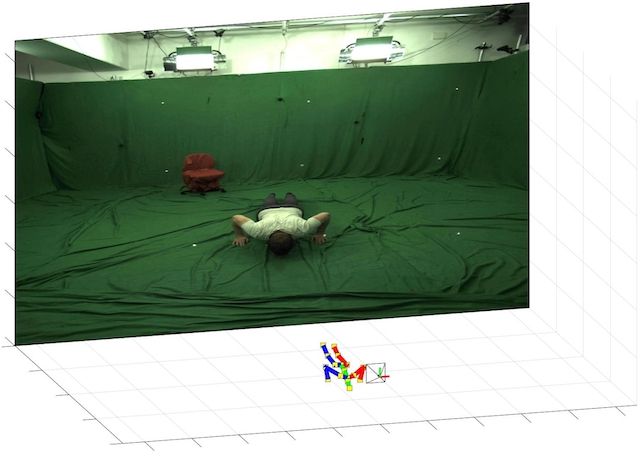}
\includegraphics[width=.46\linewidth]{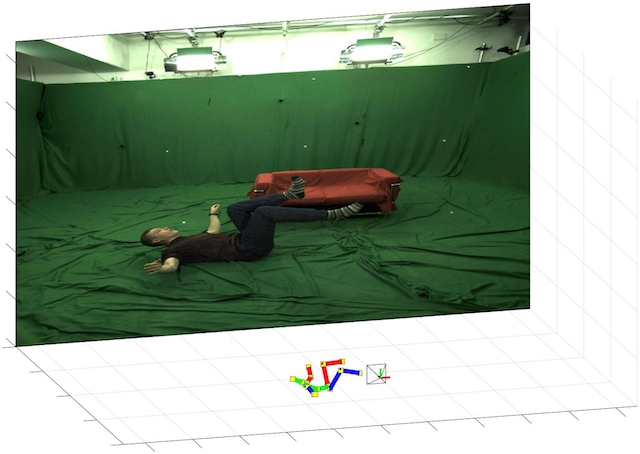}
\includegraphics[width=.46\linewidth]{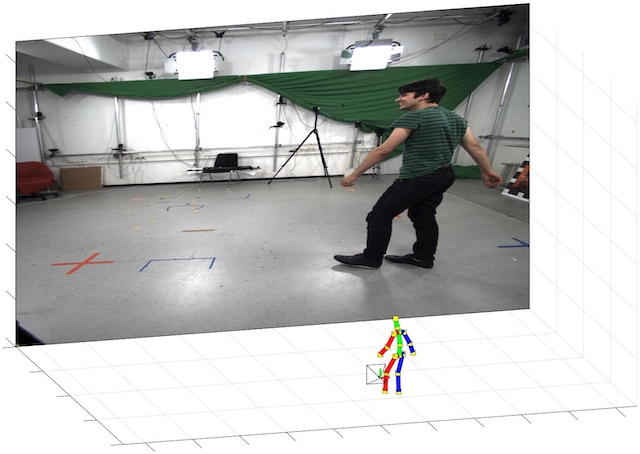}
\includegraphics[width=.46\linewidth]{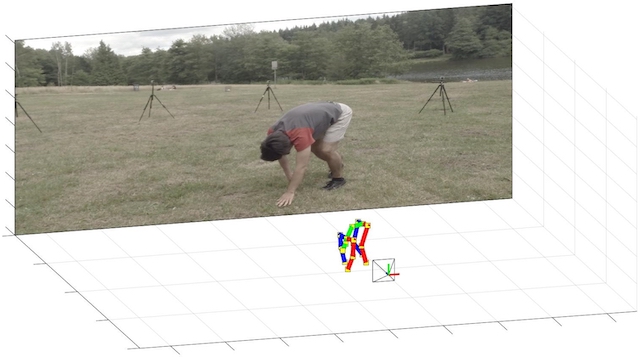}
\includegraphics[width=.46\linewidth]{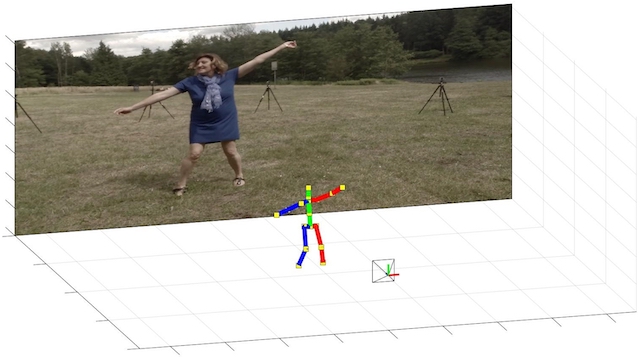}
\includegraphics[width=.46\linewidth]{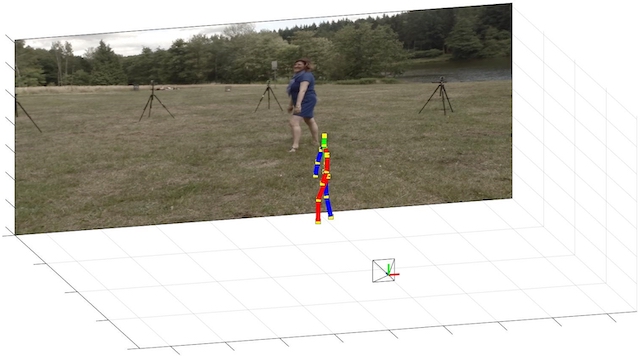}
\caption{Qualitative results of our method are shown for the MPI-INF-3DHP dataset.}
\label{fig8}
\end{figure*}

Table~\ref{table6} also presents the quantitative comparison between the proposed approach and the existing state-of-the-art methods. For the experiments using the Human3.6M dataset, our method performs better than recent state-of-the-art methods based on a geometric constraint~\cite{Zhou2017} and adversarial learning~\cite{Yang2018}. This finding shows that the proposed model can be generalized to unseen test data not used for learning. When the MPI-INF-3DHP dataset is used for learning, the proposed method outperforms all the other approaches. Figure~\ref{fig8} shows the qualitative results for some test images.

\begin{figure*}[t]
\centering
\includegraphics[width=.92\linewidth]{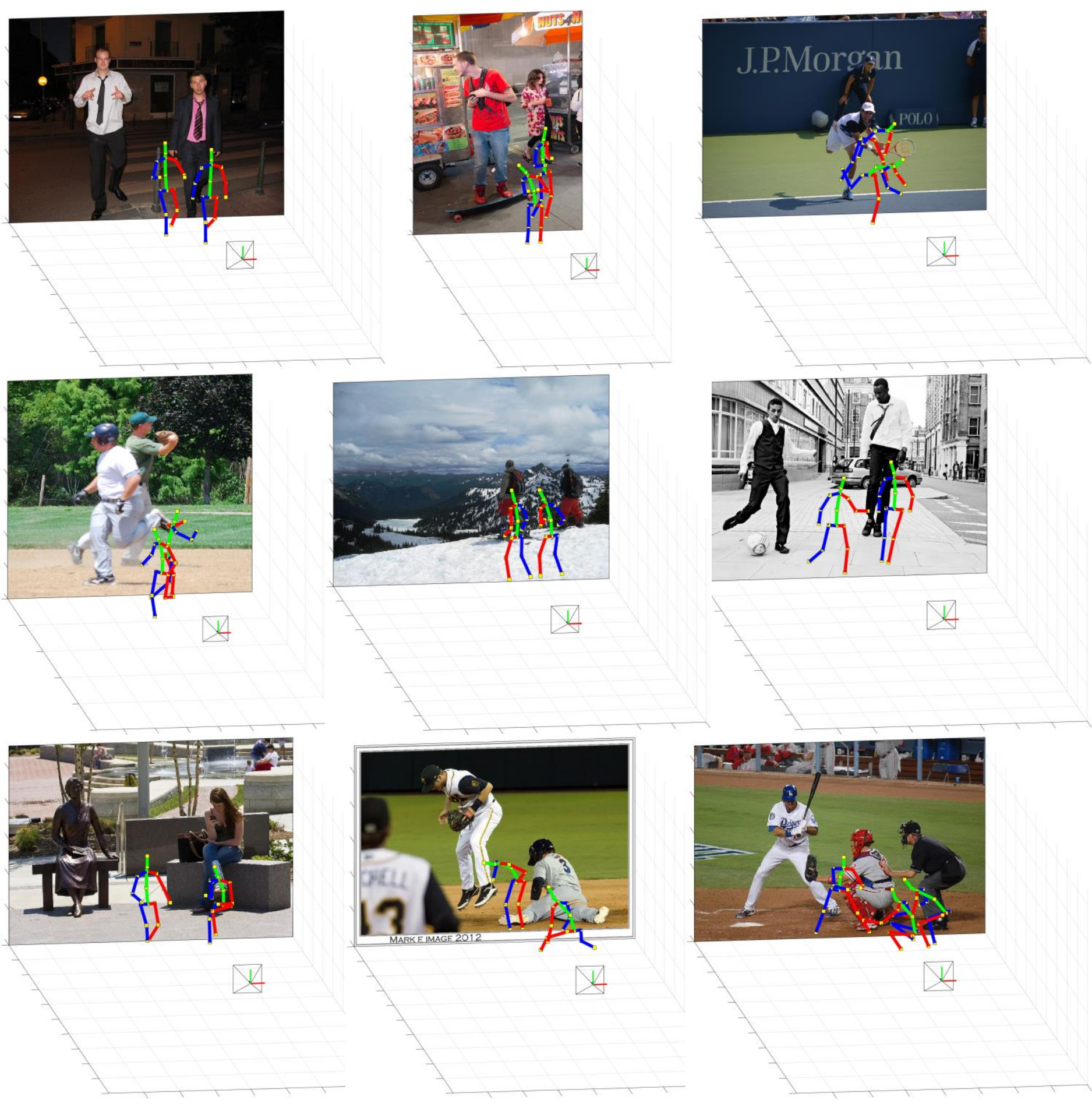}
\caption{Qualitative results of our method are shown for the in-the-wild images of the COCO dataset.}
\label{fig9}
\end{figure*}

\textbf{Qualitative results for in-the-wild images.} The proposed method is applied to the in-the-wild images of the COCO dataset~\cite{Lin2014}. To detect bounding boxes that contain persons from an input image, we use Mask R-CNN~\cite{He2017}, which is pre-trained from the COCO dataset. We then estimate the absolute 3D pose by feeding the image and detected bounding boxes to our cascade model. The focal length is manually selected. Figure~\ref{fig9} shows the estimated 3D poses. The proposed method can be used in conjunction with the object detector to perform successful 3D pose estimation on challenging in-the-wild images.

\section{Conclusion}
\label{sec6}

In this study, we propose a novel pose-lifting method (\textit{i.e.}, PoseLifter) for estimating a 3D human pose from a 2D human pose. Unlike previous methods, the proposed method enables the acquisition of an absolute pose (\textit{i.e.}, a root-relative pose with absolute root coordinates) in a camera coordinate system and achieves state-of-the-art pose lifting performance. We additionally propose a simple cascade approach, that is, a sequential combination of a 2D pose estimator and PoseLifter, for 3D human pose estimation from a single RGB image. In this case, utilizing the error statistics of 2D pose estimation for PoseLifter learning is essential and contributes to the state-of-the-art 3D pose estimation performance of the proposed approach.

{\small
\bibliographystyle{ieee}
\bibliography{mybib_2020}
}

\end{document}